\documentclass[review]{elsarticle}

\usepackage{amsmath,amsfonts,amssymb}
\usepackage{booktabs}
\usepackage{soul,color}
\usepackage{multirow,multicol}
\usepackage{algorithm}
\usepackage[noend]{algpseudocode}
\usepackage{afterpage}

\usepackage{hyperref}
\usepackage[capitalize]{cleveref}

\let\cite\undefined 

\author[ipn]{N. Majumder}
\author[nus]{D. Hazarika}
\author[ipn]{\\A. Gelbukh}
\author[ntu]{E. Cambria}
\author[ntu]{S. Poria}

\address[ipn]{Centro de Investigaci\'on en Computaci\'on, Instituto Polit\'ecnico Nacional, Mexico}
\address[nus]{School of Computing, National University of Singapore, Singapore}
\address[ntu]{School of Computer Science and Engineering, Nanyang Technological University, Singapore}

\begin{document}
\title{Multimodal~Sentiment~Analysis using Hierarchical~Fusion with Context~Modeling}

\begin{abstract}
    Multimodal sentiment analysis is a very actively growing field of
    research. A~pro\-mi\-sing area of opportunity in this field is to
    improve the multimodal fusion mechanism. We present a novel
    feature fusion strategy that proceeds in a hierarchical fashion, first
    fusing the modalities two in two and only then fusing all three
    modalities. On multimodal sentiment analysis of individual utterances, 
    our strategy outperforms conventional concatenation of
    features by 1\%, which amounts to 5\% reduction in error rate. 
    On utterance-level multimodal sentiment analysis of multi-utterance video clips,
    for which current state-of-the-art techniques incorporate contextual information 
    from other utterances of the same clip, our hierarchical fusion gives up to 2.4\% 
    (almost 10\% error rate reduction) over currently used concatenation.
    The implementation of our method is publicly available in the form of open-source code.
\end{abstract}

\maketitle

\section{Introduction}

On numerous social media platforms, such as YouTube, Facebook, or Instagram, people share their
opinions on all kinds of topics in the form of posts, images, and video clips. 
With the proliferation of smartphones and tablets, which has greatly boosted content sharing, 
people increasingly share their opinions on newly released products or on other topics in form of video reviews or comments. 
This is an excellent opportunity for large companies to capitalize on, by extracting
user sentiment, suggestions, and complaints on their products from these video reviews.
This information also opens new horizons to improving our quality of life by making informed
decisions on the choice of products we buy, services we use, places we visit, or movies we watch
basing on the experience and opinions of other users.

Videos convey information through three channels: audio, video, and text (in
the form of speech). Mining opinions from this plethora of multimodal data calls for
a solid multimodal sentiment analysis technology. One of the major problems
faced in multimodal sentiment analysis is the fusion of features pertaining to
different modalities. For this, the majority of the recent works in multimodal sentiment
analysis have simply concatenated the feature vectors of different
modalities. However, this does not take into account that different modalities
may carry conflicting information. We hypothesize that the fusion method we
present in this paper deals with this issue better, and present experimental
evidence showing improvement over simple concatenation of feature vectors. Also,
following the state of the art~\citep{porcon}, we employ recurrent
neural network (RNN) to propagate contextual information between utterances in a
video clip, which significantly improves the classification results and outperforms the
state of the art by a significant margin of 1--2\% for all the modality
combinations.

In our method, we first obtain unimodal features for each utterance for all
three modalities. Then, using RNN we extract context-aware utterance features. 
Thus, we transform the context-aware utterance vectors to the
vectors of the same dimensionality. We assume that these transformed vectors contain
abstract features representing the attributes relevant to sentiment
classification. Next, we compare and combine each bimodal combination of these
abstract features using fully-connected layers. This yields fused bimodal
feature vectors. Similarly to the unimodal case, we use RNN to generate
context-aware features. Finally, we combine these bimodal vectors into a
trimodal vector using, again, fully-connected layers and use a RNN to pass
contextual information between them. We empirically show that the feature vectors
obtained in this manner are more useful for the sentiment classification task.

The implementation of our method is publicly available in the form of open-source code.\footnote{\url{http://github.com/senticnet}}

This paper is structured as follows: \cref{sec:related-work-1} briefly discusses important previous work in multimodal feature fusion; \cref{sec:model} describes our method in details; \cref{sec:experiments} reports the results of our experiments and discuss their implications; finally, \cref{sec:conclusions} concludes the paper and discusses future work.

\section{Related Work}
\label{sec:related-work-1}
In recent years, sentiment analysis has become increasingly popular for processing social media data on online communities, blogs, wikis, microblogging platforms, and other online collaborative media~\citep{camacsa}. Sentiment analysis is a branch of affective computing research~\citep{porrev} that aims to classify text -- but sometimes also audio and video~\citep{hazcon} -- into either positive or negative -- but sometimes also neutral~\citep{chadis}. Most of the literature is on English language but recently an increasing number of works are tackling the multilinguality issue~\citep{loomul,dashtipour2016multilingual}, especially in booming online languages such as Chinese~\citep{penlea}.
Sentiment analysis techniques can be broadly categorized into symbolic and sub-symbolic approaches: the former include the use of lexicons~\citep{banlex}, ontologies~\citep{draont}, and semantic networks~\citep{camnt5} to encode the polarity associated with words and multiword expressions; the latter consist of supervised~\citep{onesta}, semi-supervised~\citep{hussem} and unsupervised~\citep{liilea} machine learning techniques that perform sentiment classification based on word co-occurrence frequencies. Among these, the most popular recently are algorithms based on deep neural networks~\citep{yourec} and generative adversarial networks~\citep{liigen}.

While most works approach it as a simple categorization problem, sentiment analysis is actually a suitcase research problem~\citep{camsui} that requires tackling many NLP tasks, including word polarity disambiguation~\citep{xiawor}, subjectivity detection~\citep{chasub}, personality recognition~\citep{majdee}, microtext normalization~\citep{satpho}, concept extraction~\citep{dhegra}, time tagging~\citep{zhotem}, and aspect extraction~\citep{maatar}.

Sentiment analysis has raised growing interest both within the scientific community, leading to many exciting open challenges, as well as in the business world, due to the remarkable benefits to be had from financial~\citep{xinfin} and political~\citep{ebrcha} forecasting, e-health~\citep{campat} and e-tourism~\citep{valsen}, user profiling~\citep{mihwha} and community detection~\citep{cavlea}, manufacturing and supply chain applications~\citep{xuuada}, human communication comprehension~\citep{zadatt} and dialogue systems~\citep{youaug}, etc.

In the field of emotion recognition, early works by~\citet{de1997facial} and
\citet{chen1998multimodal} showed that fusion of audio and visual
systems, creating a bimodal signal, yielded a higher accuracy than any unimodal
system. Such fusion has been analyzed at both feature
level~\citep{kessous2010multimodal} and decision
level~\citep{schuller2011recognizing}.

Although there is much work done on audio-visual fusion for emotion recognition,
exploring contribution of text along with audio and visual modalities in
multimodal emotion detection has been little
explored. \citet{wollmer2013youtube} and~\citet{rozgic2012ensemble} fused
information from audio, visual and textual modalities to extract emotion and
sentiment. \citet{metallinou2008audio} and~\citet{eyben2010line} fused audio and
textual modalities for emotion recognition. Both approaches relied on a
feature-level fusion. \citet{wu2011emotion} fused audio and textual clues at
decision level. \citet{pordee} uses convolutional neural network (CNN) to
extract features from the modalities and then employs multiple-kernel learning
(MKL) for sentiment analysis. The current state of the art, set forth by
\citet{porcon}, extracts contextual information from the surrounding utterances
using long short-term memory (LSTM). \citet{porrev} fuses different
modalities with deep learning-based tools. \citet{zadten} uses tensor
fusion. \citet{porens} further extends upon the ensemble of CNN and MKL.

Unlike existing approaches, which use simple concatenation based early fusion~\citep{pordee,pordep} and non-trainable tensors based fusion~\citep{zadten}, this work proposes a hierarchical fusion capable of learning the bimodal and trimodal correlations for data fusion using deep neural networks. The method is end-to-end and, in order to accomplish the fusion, it can be plugged into any deep neural network based multimodal sentiment analysis framework. 

\section{Our Method}
\label{sec:model}

In this section, we discuss our novel methodology behind solving the sentiment
classification problem. First we discuss the overview of our method and then we
discuss the whole method in details, step by step.

\subsection{Overview}
\label{sec:overview}

\subsubsection{Unimodal Feature Extraction}
We extract utterance-level features for three modalities. This step is discussed
in \cref{UFE}.

\subsubsection{Multimodal Fusion}

\begin{figure}[ht]
    \centering
    \includegraphics[scale=0.46]{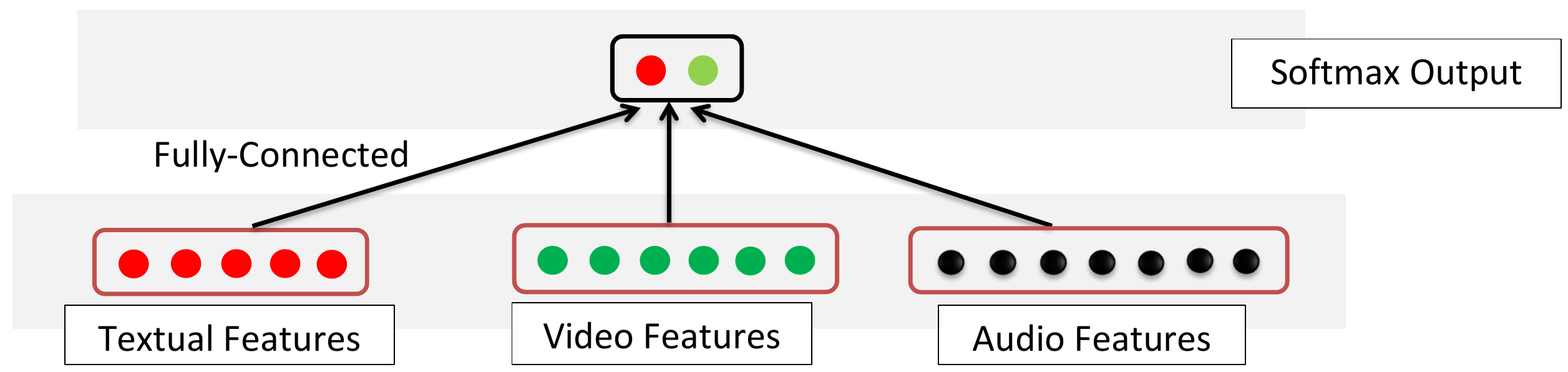}
    \caption{Utterance-level early fusion, or simple concatenation}
    \label{fig:early_fusion}
\end{figure}

\paragraph{Problems of early fusion}
The majority of the work on multimodal data use concatenation, or early fusion
(\cref{fig:early_fusion}), as their fusion strategy. The problem with this
simplistic approach is that it cannot filter out and conflicting or redundant
information obtained from different modalities. To address this major issue, we
devise an hierarchical approach which proceeds from unimodal to bimodal vectors
and then bimodal to trimodal vectors.

\begin{figure}[ht]
    \centering
    \includegraphics[scale=0.46]{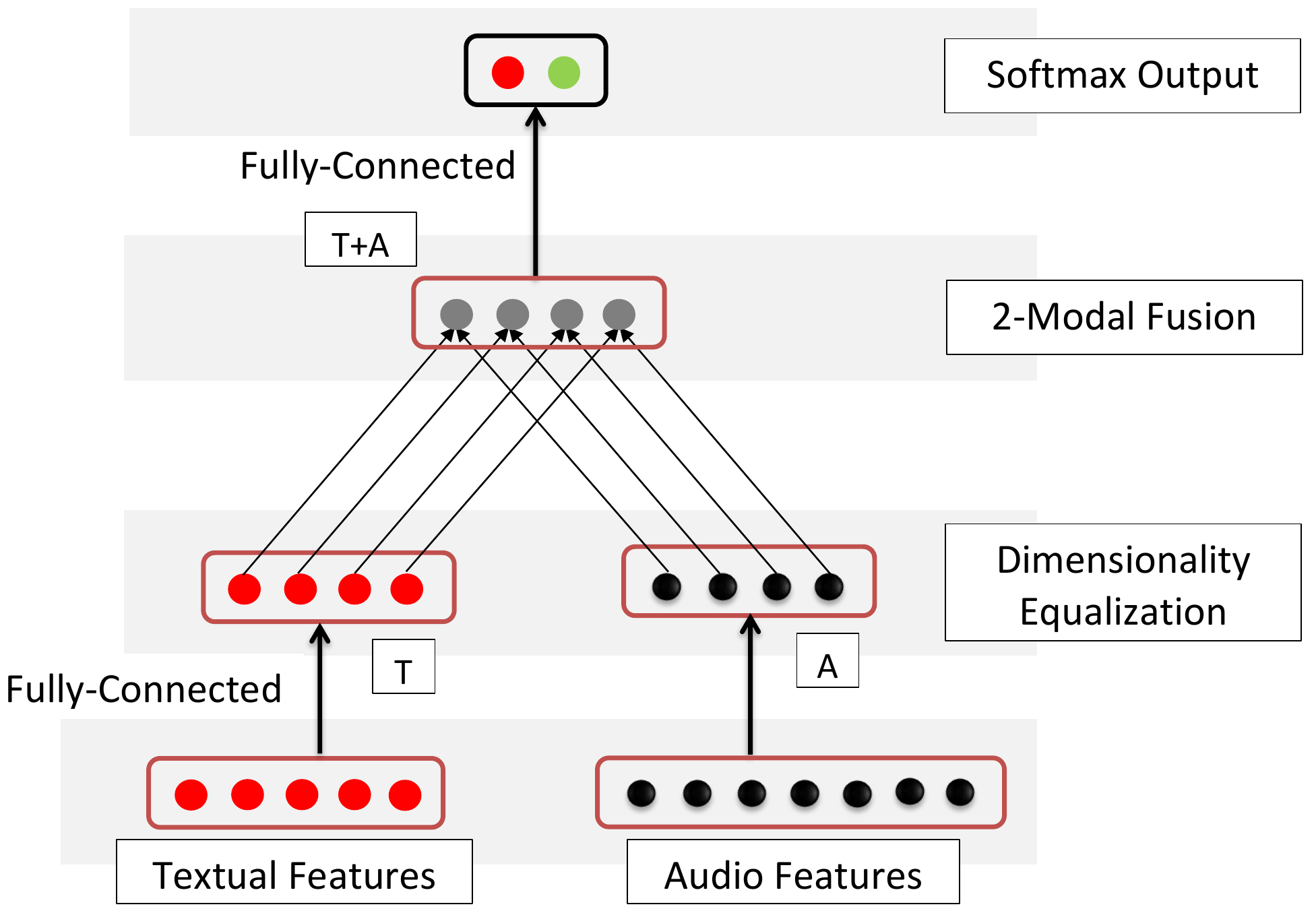}
    \caption{Utterance-level bimodal fusion}
    \label{fig:hfusion-bimodal}
\end{figure}

\paragraph{Bimodal fusion}
We fuse the utterance feature vectors for each bimodal combination, i.e., T+V,
T+A, and A+V. This step is depicted in \cref{fig:hfusion-bimodal} and discussed
in details in \cref{sec:bimodal}.
We use the penultimate layer for \cref{fig:hfusion-bimodal} as bimodal features.

\begin{figure}[ht]
    \centering
    \includegraphics[scale=0.46]{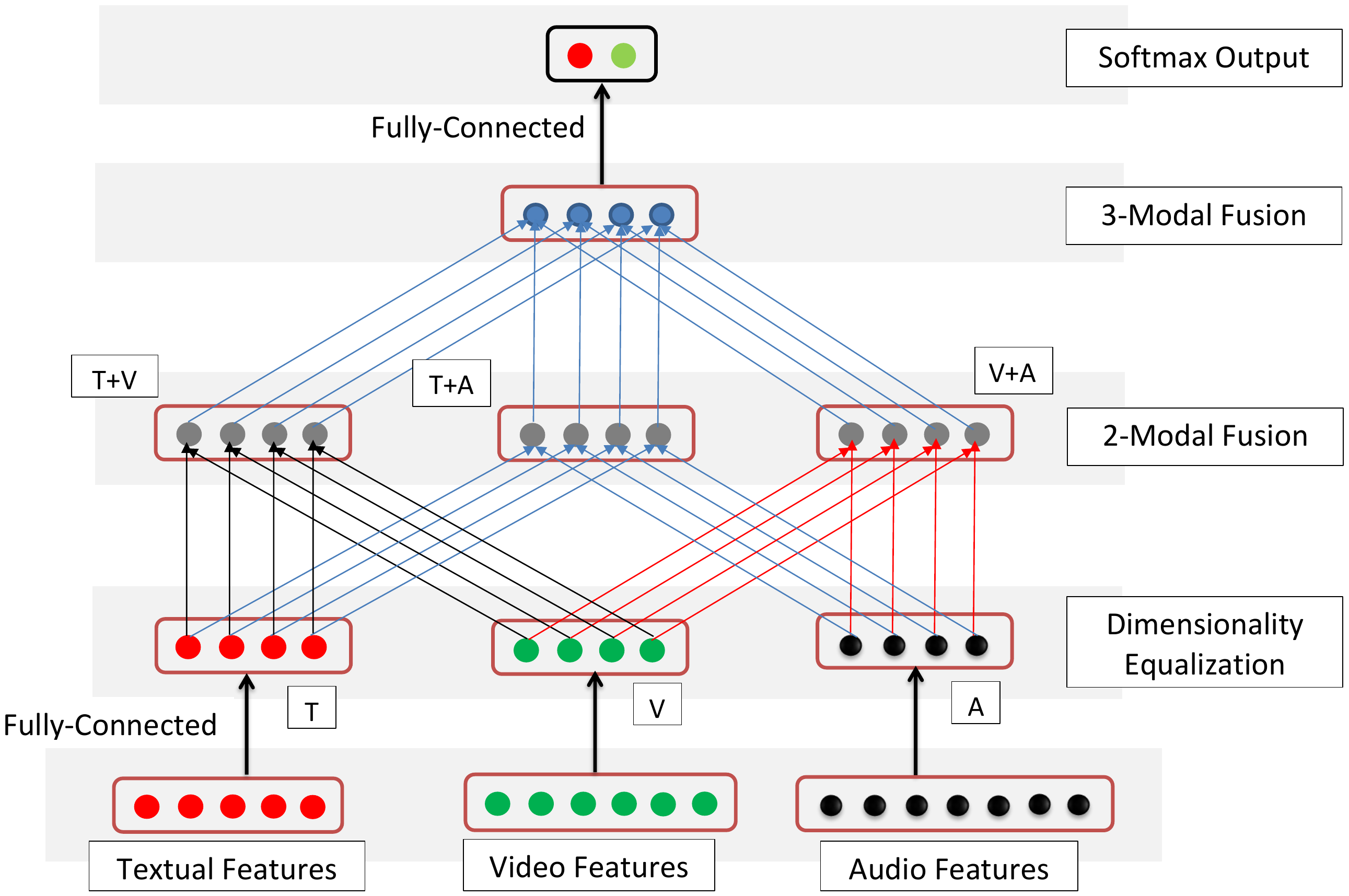}
    \caption{Utterance-level trimodal hierarchical fusion.\protect\footnotemark}
    \label{fig:hfusion-trimodal}
\end{figure}

\paragraph{Trimodal fusion}
We fuse the three bimodal features to obtain trimodal feature as depicted in
\cref{fig:hfusion-trimodal}. This step is discussed in details in \cref{sec:trimodal}.

\begin{figure}[ht]
    \centering
    \includegraphics[width=\textwidth]{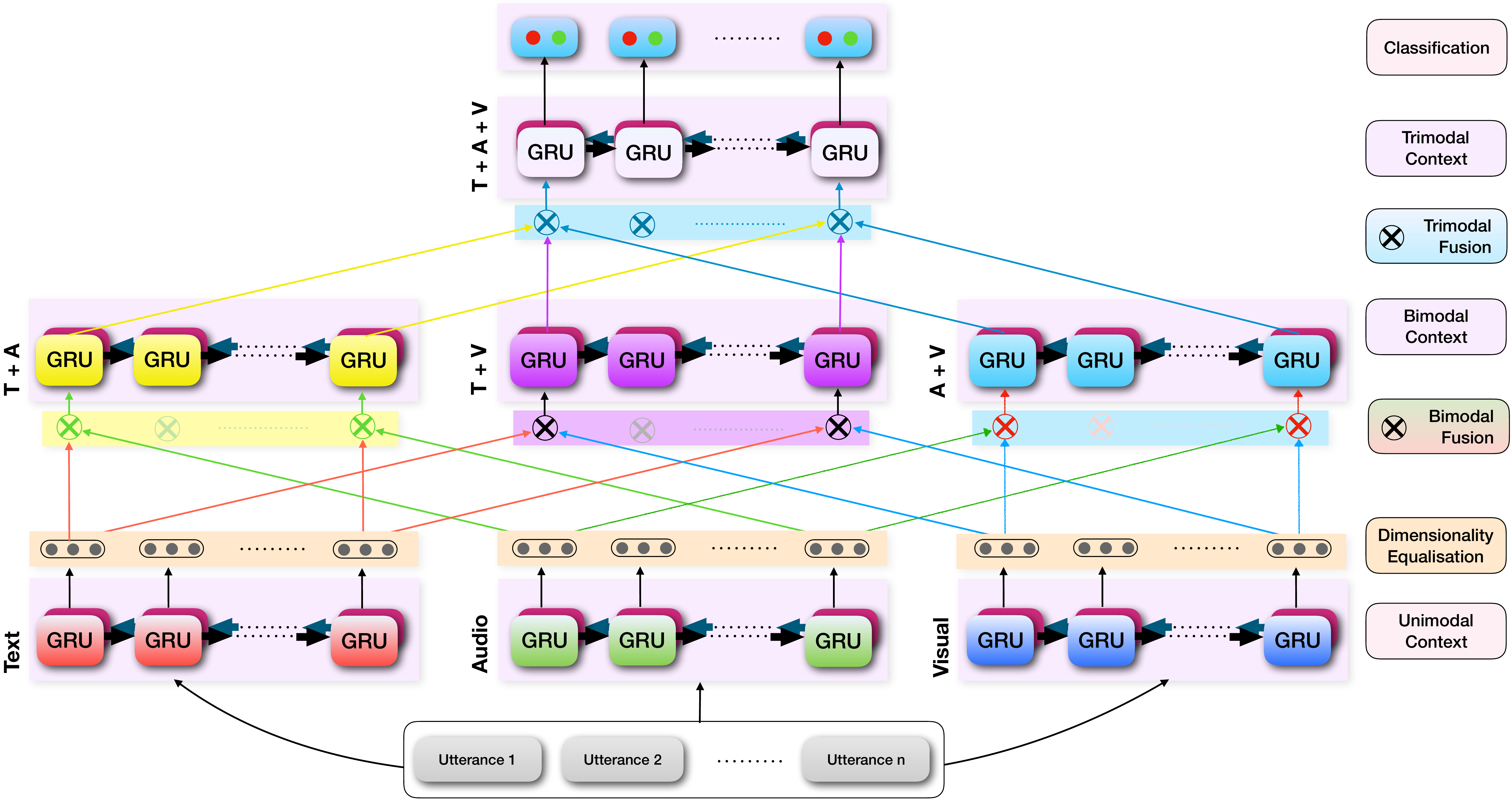}
    \caption{Context-aware hierarchical fusion}
    \label{fig:architecture}
\end{figure}

\paragraph{Addition of context}
We also improve the quality of feature vectors (both unimodal and multimodal) by
incorporating information from surrounding utterances using RNN. We model the
context using gated recurrent unit (GRU) as depicted in \cref{fig:architecture}.
The details of context modeling is discussed in \cref{sec:context} and the
following subsections.

\paragraph{Classification}
We classify the feature vectors using a softmax layer.

\subsection{Unimodal Feature Extraction}
\label{UFE}

In this section, we discuss the method of feature extraction for three different
modalities: audio, video, and text.

\subsubsection{Textual Feature Extraction}
\label{text}

The textual data is obtained from the transcripts of the videos. We apply a deep
Convolutional Neural Networks (CNN)~\citep{karpathy2014large} on each utterance
to extract textual features. Each utterance in the text is represented as an
array of pre-trained 300-dimensional {\tt word2vec}
vectors~\citep{mikolov2013efficient}. Further, the utterances are truncated or
padded with null vectors to have exactly 50 words.

Next, these utterances as array of vectors are passed through two different
convolutional layers; first layer having two filters of size 3 and 4
respectively with 50 feature maps each and the second layer has a filter of size
2 with 100 feature maps. Each convolutional layer is followed by a max-pooling
layer with window $2\times 2$.

The output of the second max-pooling layer is fed to a fully-connected layer
with 500 neurons with a rectified linear unit (ReLU)~\citep{whyeteh2001rate}
activation, followed by softmax output. The output of the penultimate
fully-connected layer is used as the textual feature. The translation of
convolution filter over makes the CNN learn abstract features and with each
subsequent layer the context of the features expands further.




\subsubsection{Audio Feature Extraction}
\label{audio}

The audio feature extraction process is performed at 30 Hz frame rate with 100
ms sliding window. We use openSMILE~\citep{eyben2010opensmile}, which is capable
of automatic pitch and voice intensity extraction, for audio feature
extraction. Prior to feature extraction audio signals are processed with voice
intensity thresholding and voice normalization. Specifically, we use
Z-standardization for voice normalization. In order to filter out audio segments
without voice, we threshold voice intensity. OpenSMILE is used to perform both
these steps. Using openSMILE we extract several Low Level Descriptors (LLD)
(e.g., pitch , voice intensity) and various statistical functionals of them
(e.g., amplitude mean, arithmetic mean, root quadratic mean, standard deviation,
flatness, skewness, kurtosis, quartiles, inter-quartile ranges, and linear
regression slope). ``IS13-ComParE'' configuration file of openSMILE is used to
for our purposes. Finally, we extracted total 6392 features from each input
audio segment.



\subsubsection{Visual Feature Extraction}
\label{visual}

To extract visual features, we focus not only on feature extraction from each
video frame but also try to model temporal features across frames. To achieve
this, we use 3D-CNN on the video. 3D-CNNs have been successful in the past,
specially in the field of object classification on 3D data~\citep{ji20133d}. Its
state-of-the-art performance on such tasks motivates its use in this paper.

Let the video be called $vid \in \mathbb{R}^{3\times f\times h\times w}$, where
$3$ represents the three RGB channels of an image and $f,\ h,\text{ and }w$
denote the cardinality, height, and width of the frames, respectively. A 3D
convolutional filter, named $f_{lt}\in \mathbb{R}^{f_m\times 3\times f_d\times
f_h\times f_w}$, is applied to this video, where, similar to a 2D-CNN, the
filter translates across the video and generates the convolution output
$conv_{out} \in \mathbb{R}^{f_m\times 3\times (f-f_d+1)\times (h-f_h+1)\times
(w-f_w+1)}$. Here, $f_m,\ f_d,\ f_h,\text{ and }f_w$ denote number of feature
maps, depth of filter, height of filter, and width of filter,
respectively. Finally, we apply max-pooling operation to the $conv_{out}$, which
selects the most relevant features. This operation is applied only to the last
three dimensions of $conv_{out}$. This is followed by a dense layer and softmax
computation. The activations of this layer is used as the overall video features
for each utterance video.

In our experiments, we receive the best results with filter dimensions $f_m =
32$ and $f_d,f_h,f_w = 5$. Also, for the max-pooling, we set the window size as
$3\times 3\times 3$ and the succeeding dense layer with $300$ neurons.

\footnotetext{Figure adapted from~\citep{mastersthesis} with permission.}

\subsection{Context Modeling}
\label{sec:context}

Utterances in the videos are semantically dependent on each other. In other
words, complete meaning of an utterance may be determined by taking preceding
utterances into consideration. We call this the context of an utterance.
Following~\citet{porcon}, we use RNN, specifically GRU\footnote{LSTM does not
  perform well} to model semantic
dependency among the utterances in a video.

Let the following items represent unimodal features:
\begin{align*}
    f_A \in \mathbb{R}^{N\times d_A}&\quad\text{(acoustic features)},\\
    f_V \in \mathbb{R}^{N\times d_V}&\quad\text{(visual features)},\\
    f_T \in \mathbb{R}^{N\times d_T}&\quad\text{(textual features)},
\end{align*}
where $N=$ maximum number of utterances in a video. We pad the shorter videos
with dummy utterances represented by null vectors of corresponding length.
For each modality, we feed the unimodal utterance features $f_m$ (where $m \in
\{A,V,T\}$) (discussed in \cref{UFE}) of a video to $GRU_m$ with
output size $D_m$, which is defined as
\begin{flalign*}
    z_m&=\sigma(f_{mt}U^{mz}+s_{m(t-1)}W^{mz}),\\
    r_m&=\sigma(f_{mt}U^{mr}+s_{m(t-1)}W^{mr}),\\
    h_{mt}&=\tanh(f_{mt}U^{mh}+(s_{m(t-1)}*r_m)W^{mh}),\\
    F_{mt}&=\tanh(h_{mt}U^{mx}+u^{mx}),\\
    s_{mt}&=(1-z_m)*F_{mt}+z_m*s_{m(t-1)},
\end{flalign*}
where $U^{mz} \in \mathbb{R}^{d_m\times D_m}$, $W^{mz} \in \mathbb{R}^{D_m\times
D_m}$, $U^{mr} \in \mathbb{R}^{d_m\times D_m}$, $W^{mr} \in
\mathbb{R}^{D_m\times D_m}$, $U^{mh} \in \mathbb{R}^{d_m\times D_m}$, $W^{mh}
\in \mathbb{R}^{D_m\times D_m}$, $U^{mx} \in \mathbb{R}^{d_m\times D_m}$,
$u^{mx} \in \mathbb{R}^{D_m}$, $z_m \in \mathbb{R}^{D_m}$, $r_m \in
\mathbb{R}^{D_m}$, $h_{mt} \in \mathbb{R}^{D_m}$, $F_{mt} \in \mathbb{R}^{D_m}$,
and $s_{mt} \in \mathbb{R}^{D_m}$. This yields hidden outputs $F_{mt}$ as
context-aware unimodal features for each modality. Hence, we define
$F_m=GRU_m(f_m)$, where $F_m \in \mathbb{R}^{N\times D_m}$. Thus, the
context-aware multimodal features can be defined as
\begin{flalign*}
    F_A &= GRU_A(f_A),\\
    F_V &= GRU_V(f_V),\\
    F_T &= GRU_T(f_T).
\end{flalign*}

\subsection{Multimodal Fusion}
\label{sec:mul_fusion}

In this section, we use context-aware unimodal features $F_A, F_V,$ and $F_T$ to
a unified feature space.

The unimodal features may have different dimensions, i.e., $D_A\neq D_V\neq
D_T$. Thus, we map them to the same dimension, say $D$ (we obtained best
results with $D=400$), using fully-connected layer as follows:
\begin{flalign*}
    g_A &= \tanh(F_AW_A+b_A),\\
    g_V &= \tanh(F_VW_V+b_V),\\
    g_T &= \tanh(F_TW_T+b_T),
\end{flalign*}
where $W_A \in \mathbb{R}^{D_A \times D}$, $b_A\in \mathbb{R}^D$, $W_V \in
\mathbb{R}^{D_V\times D}$, $b_V\in \mathbb{R}^D$, $W_T \in
\mathbb{R}^{D_T\times D}$, and $b_T\in \mathbb{R}^D$. We can represent
the mapping for each dimension as
\[
    g_x=\left[
        \begin{array}{ccccc}
        c^x_{11} & c^x_{21}  & c^x_{31}  & \cdots  & c^x_{D1}\\
        c^x_{12} & c^x_{22}  & c^x_{32}  & \cdots  & c^x_{D2}\\
        \vdots   & \vdots  & \vdots  & \cdots  & \vdots\\
        c^x_{1N} & c^x_{2N}  & c^x_{3N}  & \cdots  & c^x_{DN}\\
        \end{array}
\right],
\]
where $x \in \{V,A,T\}$ and $c^x_{lt}$ are scalars for all $l=1,2,\dots,D$ and
$t=1,2,\dots,N$. Also, in $g_x$ the rows represent the utterances and the
columns the feature values. We can see these values $c^x_{lt}$ as more abstract
feature values derived from fundamental feature values (which are the components
of $f_A$, $f_V$, and $f_T$). For example, an abstract feature can be the
angriness of a speaker in a video. We can infer the degree of angriness from
visual features ($f_V$; facial muscle movements), acoustic features ($f_A$,
such as pitch and raised voice), or textual features ($f_T$, such as the language and choice of
words). Therefore, the degree of angriness can be represented by $c^x_{lt}$,
where $x$ is $A$, $V$, or $T$, $l$ is some fixed integer between $1$ and $D$, and $t$ is some
fixed integer between $1$ and $N$.

Now, the evaluation of abstract feature values from all the modalities may not
have the same merit or may even contradict each other. Hence, we need the network
to make comparison among the feature values derived from different modalities to
make a more refined evaluation of the degree of anger. To this end, we take
each bimodal combination (which are audio--video, audio--text, and video--text)  at
a time and compare and combine each of their respective abstract feature values
(i.e. $c^V_{lt}$ with $c^T_{lt}$, $c^V_{lt}$ with $c^A_{lt}$, and $c^A_{lt}$
with $c^T_{lt}$) using fully-connected layers as follows:
\begin{align}
    i^{VA}_{lt}&=\tanh(w^{VA}_{l}.[c_{lt}^V,c_{lt}^A]^\intercal+b^{VA}_{l}),\label{bimodal:1}\\
    i^{AT}_{lt}&=\tanh(w^{AT}_{l}.[c_{lt}^A,c_{lt}^T]^\intercal+b^{AT}_{l}),\label{bimodal:2}\\
    i^{VT}_{lt}&=\tanh(w^{VT}_{l}.[c_{lt}^V,c_{lt}^T]^\intercal+b^{VT}_{l}),\label{bimodal:3}   
\end{align}
where $w^{VA}_l \in \mathbb{R}^2$, $b^{VA}_l$ is scalar, $w^{AT}_l \in
\mathbb{R}^2$, $b^{AT}_l$ is scalar, $w^{VT}_l \in \mathbb{R}^2$, and $b^{VT}_l$
is scalar, for all $l=1,2,\dots,D$ and $t=1,2,\dots,N$. We hypothesize that it
will enable the network to compare the decisions from each modality against the
others and help achieve a better fusion of modalities.

\paragraph{\textbf{Bimodal fusion}}
\label{sec:bimodal}

\crefrange{bimodal:1}{bimodal:3} are used for bimodal fusion. The bimodal
fused features for video--audio, audio--text, video--text are defined as
\begin{flalign*}
    f_{VA}= (f_{VA1},f_{VA2},\dots,f_{VA(N)}), \text{ where } f_{VAt}&=(i^{VA}_{1t},i^{VA}_{2t},\dots,i^{VA}_{Dt}), \\
    f_{AT}= (f_{AT1},f_{AT2},\dots,f_{AT(N)}), \text{ where } f_{ATt}&=(i^{AT}_{1t},i^{AT}_{2t},\dots,i^{AT}_{Dt}), \\
    f_{VT}= (f_{VT1},f_{VT2},\dots,f_{VT(N)}), \text{ where } f_{VTt}&=(i^{VT}_{1t},i^{VT}_{2t},\dots,i^{VT}_{Dt}).
\end{flalign*}

We further employ $GRU_m$(~\cref{sec:context}) ($m \in \{VA, VT, TA\}$), to
incorporate contextual information among the utterances in a video with
\begin{flalign*}
    F_{VA} = (F_{VA1},F_{VA2},\dots,F_{VA(N)}) = GRU_{VA}(f_{VA}),\\
    F_{VT} = (F_{VT1},F_{VT2},\dots,F_{VT(N)}) = GRU_{VT}(f_{VT}),\\
    F_{TA} = (F_{TA1},F_{TA2},\dots,F_{TA(N)}) = GRU_{TA}(f_{TA}),
\end{flalign*}
where
\begin{flalign*}
    F_{VAt}= (I^{VA}_{1t},I^{VA}_{2t},\dots,I^{VA}_{D_2t}),\\
    F_{VTt}= (I^{AT}_{1t},I^{AT}_{2t},\dots,I^{AT}_{D_2t}),\\
    F_{TAt}= (I^{VT}_{1t},I^{VT}_{2t},\dots,I^{VT}_{D_2t}),
\end{flalign*}
$F_{VA}$, $F_{VT}$, and $F_{TA}$ are context-aware bimodal features
represented as vectors and $I^m_{nt}$ is scalar for $n=1,2,\dots,D_2$,
$D_2=500$, $t=1,2,\dots,N$, and $m=\text{VA,VT,TA}$.

\paragraph{Trimodal fusion}
\label{sec:trimodal}

We combine all three modalities using fully-connected layers as follows:
\begin{flalign*}
    z_{lt}=\tanh(w^{AVT}_l.[I^{VA}_{lt},I^{AT}_{lt},I^{VT}_{lt}]^\intercal+b^{AVT}_l),
\end{flalign*}
where $w^{AVT}_l \in \mathbb{R}^3$ and $b^{AVT}_l$ is a scalar for all $l=1,2,\dots,D_2$
and $t=1,2,\dots,N$.
So, we define the fused features as
\begin{flalign*}
    f_{AVT}=(f_{AVT1},f_{AVT2},\dots,f_{AVT(N)}),
\end{flalign*}
where
$f_{AVTt}=(z_{1t},z_{2t},\dots,z_{D_2t})$,
$z_{nt}$ is scalar for $n=1,2,\dots,D_2$ and $t=1,2,\dots,N$.

Similarly to bimodal fusion (\cref{sec:bimodal}), after trimodal fusion we pass
the fused features through $GRU_{AVT}$ to incorporate contextual information in
them, which yields
\begin{flalign*}
    F_{AVT} = (F_{AVT1},F_{AVT2},\dots,F_{AVT(N)}) = GRU_{AVT}(f_{AVT}),
\end{flalign*}
where $F_{AVTt}= (Z_{1t},Z_{2t},\dots,Z_{D_3t})$, $Z_{nt}$ is scalar for $n=1,2,\dots,D_3$, $D_3=550$, $t=1,2,\dots,N$,
and $F_{AVT}$ is the context-aware trimodal feature vector.

\subsection{Classification}
\label{sec:classification}

In order to perform classification, we feed the fused features $F_{mt}$ (where
$m=AV,VT,TA,\text{ or } AVT$ and $t=1,2,\dots,N$) to a softmax layer with $C=2$
outputs. The classifier can be described as follows:
\begin{flalign*}
    \mathcal{P} &=
    \text{softmax}(W_{\mathit{softmax}}F_{mt}+b_{\mathit{softmax}}),\\
    \hat{y}&=\underset{j}{\text{argmax}}(\mathcal{P}[j]),
\end{flalign*}
where $W_{\mathit{softmax}}\in \mathbb{R}^{C\times D}$,
$b_{\mathit{softmax}}\in \mathbb{R}^C$, $\mathcal{P}\in \mathbb{R}^C$, $j=$
class value ($0$ or $1$), and $\hat{y}=$ estimated class value.

\subsection{Training}
\label{training}
We employ categorical cross-entropy as loss function ($J$) for training,
\begin{flalign*}
    J=-\frac{1}{N}\sum_{i=1}^N{\sum_{j=0}^{C-1}{y_{ij}\log{\mathcal{P}_i[j]}}},
\end{flalign*}
where $N=$ number of samples, $i=$ index of a sample, $j=$ class value, and
\[
    y_{ij}=
        \begin{cases}
            1, & \text{if expected class value of sample }i\text{ is }j\\
            0, & \text{otherwise.}
        \end{cases}
\]

  Adam~\citep{DBLP:journals/corr/KingmaB14} is used as optimizer due to its
  ability to adapt learning rate for each parameter individually. We train the
network for 200 epochs with early stopping, where we optimize the parameter set
  \begin{flalign*}
\theta=&\bigcup_{m\in M}\left (\bigcup_{j\in \{z,r,h\}}\{U^{mj},W^{mj}\}\cup \{U^{mx},u^{mx}\}\right
)\\
&\cup \bigcup_{m\in M_2}\bigcup_{i=1}^{D_2}\{w^m_i\} \cup
\bigcup_{i=1}^{D_3}\{w^{AVT}_i\}\cup \bigcup_{m\in M_1}\{W_m,b_m\}\\
&\cup \{W_{softmax},b_{softmax}\},\\
  \end{flalign*}
where $M=\{A,V,T,VA,VT,TA,AVT\}$, $M_1=\{A,V,T\}$, and
$M_2=\{VA,VT,\allowbreak TA\}$. \cref{algorithm} summarizes our method.\footnote{Implementation of this algorithm is available at
  \url{http://github.com/senticnet}} 

\begin{algorithm}[!ht]
    \small
    \caption{Context-Aware Hierarchical Fusion Algorithm}\label{algorithm}
    \begin{algorithmic}[1]

        \vspace{2mm}
        \Procedure{TrainAndTestModel}{$U$, $V$}\Comment{\footnotesize{$U$ = train set, $V$ = test set}}
        \vspace{2mm}
        \State \textbf{Unimodal feature extraction:}
        \For{\texttt{i:[1,$N$]}}\Comment{\footnotesize{extract baseline features}}
            \State \texttt{$f_{A}^{i} \gets AudioFeatures(u_{i})$ }
            \State \texttt{$f_{V}^{i} \gets VideoFeatures(u_{i})$ }
            \State \texttt{$f_{T}^{i} \gets TextFeatures(u_{i})$ }
        \EndFor
        \For{\texttt{m $\in \{A, V, T\}$}}
            \State $F_m$ = $GRU_m$($f_m$)
        \EndFor
        \vspace{2mm}
        \State \textbf{Fusion}:
        \State \texttt{$g_{A} \gets MapToSpace(F_A)$ }\Comment{\footnotesize{dimensionality equalization}}
        \State \texttt{$g_{V} \gets MapToSpace(F_V)$ }
        \State \texttt{$g_{T} \gets MapToSpace(F_T)$ }

        \vspace{2mm}
        \State \texttt{$f_{VA} \gets BimodalFusion(g_V, g_A)$}\Comment{\footnotesize{bimodal fusion}}
        \State \texttt{$f_{AT} \gets BimodalFusion(g_A, g_T)$}
        \State \texttt{$f_{VT} \gets BimodalFusion(g_V, g_T)$}
        \For{\texttt{m $\in \{VA, AT, VT\}$}}
        \State $F_m$ = $GRU_m$($f_m$)
        \EndFor
        
        \vspace{2mm}
        \State $f_{AVT} \gets TrimodalFusion(F_{VA}, F_{AT}, F_{VT})$
        \Comment{\small{trimodal fusion}}
        \State $F_{AVT}$ = $GRU_{AVT}$($f_{AVT}$)

        \vspace{2mm}
        \For{\texttt{i:[1,$N$]}}\Comment{\footnotesize{softmax classification}}
            \State $\hat{y}^i =\underset{j}{\text{argmax}}(softmax(F_{AVT}^i)[j])$ 
        \EndFor

        \State $TestModel(V)$
        \EndProcedure
        
        \vspace{2mm}
        \Procedure{MapToSpace}{$x_z$} \Comment{\footnotesize{for modality $z$}}
        \State $ g_z \gets \tanh(W_zx_z+b_z)  $
        \State \textbf{return} $g_z$
        \EndProcedure

        \vspace{2mm}
        \Procedure{BimodalFusion}{$g_{z_1}$, $g_{z_2}$} \Comment{\footnotesize{for modality
        $z_1$ and $z_2$, where $z_1\neq z_2$}}
        \For{\texttt{i:[1,$D$]}}
        \State $f_{z_1z_2}^i \gets \tanh(w_i^{z_1z_2}.[g^i_{z_1},
        g^i_{z_2}]^\intercal+b_i^{z_1z_2})$
        \EndFor
        \State $f_{z_1z_2} \gets (f_{z_1z_2}^1, f_{z_1z_2}^2,\dots,f_{z_1z_2}^{D})$
        \State \textbf{return} $f_{z_1z_2}$
        \EndProcedure

        \vspace{2mm}
        \Procedure{TrimodalFusion}{$f_{z_1}$, $f_{z_2}$, $f_{z_3}$} \Comment{\footnotesize{for
                modality combination $z_1$, $z_2$, and $z_3$, where $z_1\neq z_2\neq z_3$}}
        \For{\texttt{i:[1,$D$]}}
        \State $f^i_{z_1z_2z_3} \gets \tanh(w_i.[f^i_{z_1}, f^i_{z_2}, f^i_{z_3}]^\intercal+b_i)$
        \EndFor
        \State $f_{z_1z_2z_3} \gets (f_{z_1z_2z_3}^1, f_{z_1z_2z_3}^2,\dots,f_{z_1z_2z_3}^{D})$
        \State \textbf{return} $f_{z_1z_2z_3}$
        \EndProcedure
        
        \vspace{2mm}
        \Procedure{TestModel}{$V$}
        \State \footnotesize{Similarly to training phase, $V$ is passed through the learnt models
        to get the features and classification outputs. \cref{training}
        mentions the trainable parameters ($\theta$).}
        \EndProcedure
    \end{algorithmic}
\end{algorithm}

\section{Experiments}
\label{sec:experiments}

\subsection{Dataset Details}
\label{datasets}
Most research works in multimodal sentiment analysis are performed on datasets
where train and test splits may share certain speakers. Since, each individual
has an unique way of expressing emotions and sentiments, finding generic and
person-independent features for sentiment analysis is
crucial. \cref{tab:dataset} shows the train and test split for the datasets
used.

\begin{table}[h]
	\small
	      \addtolength\tabcolsep{-5pt} 
	\begin{center}
		\begin{tabular}{|*{14}{c|}}
			\hline
			\multicolumn{2}{|c|}{\multirow{2}{*}{Dataset}} & \multicolumn{6}{c|}{Train} & \multicolumn{6}{c|}{Test}\\ \cline{3-14}
			\multicolumn{2}{|c|}{}& \emph{pos.}&\emph{neg.}&\emph{happy}&\emph{anger}&\emph{sad}&\emph{neu.}&\emph{pos.}&\emph{neg.}&\emph{happy}&\emph{anger}&\emph{sad}&\emph{neu.}\\ \hline
			\multicolumn{2}{|c|}{MOSI}&709&738&-&-&-&-&467&285&-&-&-&-\\ \hline
			\multicolumn{2}{|c|}{IEMOCAP}&-&-&1194&933&839&1324&-&-&433&157&238&380\\ \hline
			\multicolumn{14}{l}{\scriptsize{pos. = positive, neg. = negative, neu. = neutral}}
		\end{tabular}
	\end{center}
	\vspace{-4.5mm}
	\caption {Class distribution of datasets in both train and test splits. }
	\label{tab:dataset}
\end{table}

\subsubsection{CMU-MOSI}
\label{sec:mosi}
CMU-MOSI dataset~\citep{zadeh2016multimodal} is rich in sentimental
expressions, where 89 people review various topics in English. The videos are
segmented into utterances where each utterance is annotated with scores between
$-3$ (strongly negative) and $+3$ (strongly positive) by five annotators. We
took the average of these five annotations as the sentiment polarity and
considered only two classes (positive and negative). Given every individual's
unique way of expressing sentiments, real world applications should be able to
model generic person independent features and be robust to person variance. To
this end, we perform person-independent experiments to emulate unseen
conditions. Our train/test splits of the dataset are completely disjoint with
respect to speakers. The train/validation set consists of the first 62
individuals in the dataset. The test set contains opinionated videos by rest of
the 31 speakers. In particular, 1447 and 752 utterances are used for training
and test respectively.

\subsubsection{IEMOCAP}
\label{sec:iemocap}
IEMOCAP~\citep{iemocap} contains two way conversations
among ten speakers, segmented into utterances. The utterances are tagged with
the labels anger, happiness, sadness, neutral, excitement, frustration, fear,
surprise, and other. We consider the first four ones to compare with the
state of the art~\citep{porcon} and other works. It contains 1083
angry, 1630 happy, 1083 sad, and 1683 neutral videos. Only the videos by the
first eight speakers are considered for training.

\subsection{Baselines}
We compare our method with the following strong baselines.

\paragraph{Early fusion}
\label{early-fusion}
We extract unimodal features (\cref{UFE}) and simply concatenate them to
produce multimodal features. Followed by support vector machine (SVM)
being applied on this feature vector for the final sentiment
classification.

\paragraph{Method from~\citep{pordee}}
We have implemented and compared our method with the approach proposed by
\citet{pordee}. In their approach, they extracted visual features using
CLM-Z, audio features using openSMILE, and textual features using CNN. MKL was then applied to the features obtained from
concatenation of the unimodal features. However, they did not conduct speaker
independent experiments.

In order to perform a fair comparison with~\citep{pordee}, we
employ our fusion method on the features extracted by~\citet{pordee}.


\paragraph{Method from~\citep{porcon}}
We have compared our method with~\citep{pordep}, which takes
advantage of contextual information obtained from the surrounding
utterances. This context modeling is achieved using LSTM. We reran the
experiments of~\citet{pordep} without using SVM for classification since using
SVM with neural networks is usually discouraged. This provides a fair comparison
with our model which does not use SVM.

\paragraph{Method from~\citep{zadten}}
In~\citep{zadten}, they proposed a trimodal fusion method based on the tensors. We have also compared our method with their. In particular, their dataset configuration was different than us so we have adapted their publicly available code ~\footnote{\url{https://github.com/A2Zadeh/TensorFusionNetwork}} and employed that on our dataset.
\subsection{Experimental Setting}
\label{sec:exp_set}

We considered two variants of experimental setup while evaluating our model.

\paragraph{HFusion} In this setup, we evaluated hierarchical fusion
without context-aware features with CMU-MOSI dataset. We removed all the GRUs
from the model described in \cref{sec:mul_fusion,sec:context} forwarded
utterance specific features directly to the next layer. This setup is depicted
in \cref{fig:hfusion-trimodal}. 

\paragraph{CHFusion} This setup is exactly as the model described in
\cref{sec:model}.

\subsection{Results and Discussion}

We discuss the results for the different experimental settings discussed in
\cref{sec:exp_set}.

\begin{table*}[t]
    \centering
    \small
        \caption{Comparison in terms of accuracy of Hierarchical Fusion
          (HFusion) with other fusion methods for CMU-MOSI dataset; bold font
        signifies best accuracy for the corresponding feature set and
        modality or modalities, where T stands for text, V for video, and A for audio. $SOTA^1$ = Poria et al.~\citep{pordee}, $SOTA^2$ = Zadeh et al.~\citep{zadten}}
    \resizebox{\textwidth}{!}{
      \addtolength\tabcolsep{-3pt} 
      \begin{tabular}[t]{@{\extracolsep{5pt}}ccccccc}
        \hline
        \multirow2*{\begin{tabular}{c}Modality\\ Combination\end{tabular}}&
        \multicolumn{3}{c}{\citep{pordee} feature set}&
        \multicolumn{3}{c}{Our feature set}\\
        \cline{2-4}\cline{5-7}
        & $SOTA^{1}$&
        $SOTA^2$&
        HFusion&
        Early fusion &
        $SOTA^2$&
        HFusion\\
        \cline{1-1}\cline{2-4}\cline{5-7}
        T      & \multicolumn{3}{c}{N/A}            & \multicolumn{3}{c}{75.0\%}    \\
        V      & \multicolumn{3}{c}{N/A}              & \multicolumn{3}{c}{55.3\%} \\
        A      & \multicolumn{3}{c}{N/A}              & \multicolumn{3}{c}{56.9\%} \\
        \hline
        T+V    & 73.2\% & 73.8\%  & \textbf{74.4\%}          & 77.1\% & 77.4\% & \textbf{77.8\%}\\
        T+A    & 73.2\%  & 73.5\% & \textbf{74.2\%}         & 77.1\% & 76.3\% & \textbf{77.3\%}\\
        A+V    & 55.7\%  & 56.2\% & \textbf{57.5\%}          & 56.5\% & 56.1\% & \textbf{56.8\%}\\
        \hline
        A+V+T  & 73.5\% & 71.2\% & \textbf{74.6\%} & 77.0\% & 77.3\% & \textbf{77.9\%}\\
        \hline
    \end{tabular}
    }
    \label{table:hfusion}
\end{table*}

\subsubsection{Hierarchical Fusion (HFusion)}
\label{hfusion}

The results of our experiments are presented in \cref{table:hfusion}. We
evaluated this setup with CMU-MOSI dataset (\cref{sec:mosi}) and two
feature sets: the feature set used in~\citep{pordee}
and the set of unimodal features discussed in \cref{UFE}.

Our model outperformed~\citep{pordee}, which employed MKL, for all bimodal
and trimodal scenarios by a margin of 1--1.8\%. This leads us to present two
observations. Firstly, the features used in~\citep{pordee} are inferior to
the features extracted in our approach. Second, our hierarchical
fusion method is better than their fusion method.

It is already established in the literature
\citep{pordee,perez2013utterance} that multimodal analysis outperforms
unimodal analysis. We also observe the same trend in our experiments where
trimodal and bimodal classifiers outperform unimodal classifiers. The textual
modality performed best among others with a higher unimodal classification
accuracy of 75\%. Although other modalities contribute to improve the
performance of multimodal classifiers, that contribution is little in compare to
the textual modality.

On the other hand, we compared our model with early fusion
(\cref{early-fusion}) for aforementioned feature sets
(\cref{UFE}). Our fusion mechanism consistently outperforms early fusion for
all combination of modalities. This supports our
hypothesis that our hierarchical fusion method captures the
inter-relation among the modalities and produce better performance vector than
early fusion. Text is the strongest individual modality, and we observe that
the text modality paired with remaining two modalities results in consistent
performance improvement.

Overall, the results give a strong indication that the comparison among the
abstract feature values dampens the effect of less important modalities, which
was our hypothesis. For example, we can notice that for early fusion T+V and T+A
both yield the same performance. However, with our method text with video
performs better than text with audio, which is more aligned with our
expectations, since facial muscle movements usually carry more emotional nuances
than voice.

In particular, we observe that our model outperformed all the strong baselines mentioned above. The method by~\citep{pordee} is only able to fuse using concatenation. Our proposed method outperformed their approach by a significant margin; thanks to the power of hierarchical fusion which proves the capability of our method in modeling bimodal and trimodal correlations. However on the other hand, the method by~\citep{zadten} is capable of fusing the modalities using a tensor. Interestingly our method also outperformed them and we think the reason is the capability of bimodal fusion and use that for trimodal fusion. Tensor fusion network is incapable to learn the weights of the bimodal and trimodal correlations in the fusion. Tensor Fusion is mathematically formed by an outer product, it
has no learn-able parameters. Wherein our method learns the weights automatically using a neural network (Equation 1,2 and 3).

\begin{table*}[t]
    \centering
        \caption{Comparison of Context-Aware Hierarchical Fusion (CHFusion) in terms of accuracy ($\text{CHFusion}_{acc}$) and f-score (for IEMOCAP: $\text{CHFusion}_{fsc}$) with the state of the art for CMU-MOSI
          and IEMOCAP dataset; bold font signifies best accuracy for the corresponding dataset and
          modality or modalities, where T stands text, V for video, A for audio. $SOTA^1$ = Poria et al.~\citep{pordee}, $SOTA^2$ = Zadeh et al.~\citep{zadten}. $\text{CHFusion}_{acc}$ and $\text{CHFusion}_{fsc}$ are the accuracy and f-score of CHFusion respectively.}
    \resizebox{\textwidth}{!}{
          \addtolength\tabcolsep{-6pt} 
    \begin{tabular}[t]{@{\extracolsep{4pt}}cccccccc}
      \hline
      \multirow2*{\begin{tabular}{c}Modality\end{tabular}} & \multicolumn{3}{c}{CMU-MOSI} & \multicolumn{4}{c}{IEMOCAP} \\
        \cline{2-4}\cline{5-8}
      & $SOTA^1$ & $SOTA^2$ & $\text{CHFusion}_{acc}$ & $SOTA^1$ & $SOTA^2$ & $\text{CHFusion}_{acc}$ & $\text{CHFusion}_{fsc}$\\
      \cline{1-1}\cline{2-4}\cline{5-7}\cline{8-8}
      T & \multicolumn{3}{c}{76.5\%} & \multicolumn{3}{c}{73.6\%} & -\\
      V & \multicolumn{3}{c}{54.9\%} & \multicolumn{3}{c}{53.3\%} & -\\
      A & \multicolumn{3}{c}{55.3\%} & \multicolumn{3}{c}{57.1\%} & -\\
      \hline
      T+V & 77.8\% & 77.1\% & \textbf{79.3\%} & 74.1\% & 73.7\% & \textbf{75.9\%} & 75.6\%\\
      T+A & 77.3\% & 77.0\% & \textbf{79.1\%} & 73.7\% & 71.1\% & \textbf{76.1\%} & 76.0\%\\
      A+V & 57.9\% & 56.5\% & \textbf{58.8\%} & 68.4\%  & 67.4\% & \textbf{69.5\%} & 69.6\% \\
      \hline
      A+V+T & 78.7\% & 77.2\% & \textbf{80.0\%} & 74.1\% & 73.6\% & \textbf{76.5\%} & 76.8\%\\
      \hline
    \end{tabular}
    }
    \label{table:chfusion}
\end{table*}

\subsubsection{Context-Aware Hierarchical Fusion (CHFusion)}
\label{chfusion}

The results of this experiment are shown in \cref{table:chfusion}. This setting
fully utilizes the model described in \cref{sec:model}. We applied this
experimental setting for two datasets, namely CMU-MOSI~(\cref{sec:mosi}) and
IEMOCAP~(\cref{sec:iemocap}). We used the feature set discussed in \cref{UFE},
which was also used by~\citet{porcon}. As expected our method outperformed the simple early fusion based fusion by~\citep{pordee}, tensor fusion by~\citep{zadten}. The method by~\citet{porcon} used a scheme to learn contextual features from the surrounding features. However, as a method of fusion they adapted simple concatenation based fusion method by~\citep{pordee}. As discussed in Section \ref{sec:context}, we employed their contextual feature extraction framework and integrated our proposed fusion method to that. This has helped us to outperform~\citet{porcon} by significant margin thanks to the hierarchical fusion (HFusion).

\paragraph{CMU-MOSI}
We achieve 1--2\% performance improvement over the state of the art
\citep{porcon} for all the modality combinations having textual
component. For A+V modality combination we achieve better but similar
performance to the state of the art. We suspect that it is due to both audio and
video modality being significantly less informative than textual modality. It is
evident from the unimodal performance where we observe that textual modality on
its own performs around 21\% better than both audio and video modality. Also,
audio and video modality performs close to majority baseline. On the other hand,
it is important to notice that with all modalities combined we achieve about
3.5\% higher accuracy than text alone.

For example, consider the following utterance: \emph{so overall new moon even with the bigger better budgets huh it was still too long}.
The speaker discusses her opinion on the movie Twilight New Moon. Textually the
utterance is abundant with positive words however audio and video comprises of a
frown which is observed by the hierarchical fusion based model.

\paragraph{IEMOCAP}
As the IEMOCAP dataset contains four distinct emotion categories, in the last layer of the network we used a softmax classifier whose output dimension is set to 4. 
In order to perform classification on IEMOCAP dataset we feed the fused features $F_{mt}$ (where
$m=AV,VT,TA,\text{ or } AVT$ and $t=1,2,\dots,N$) to a softmax layer with $C=4$
outputs. The classifier can be described as follows:
\begin{flalign*}
\mathcal{P} &=
\text{softmax}(W_{\mathit{softmax}}F_{mt}+b_{\mathit{softmax}}),\\
\hat{y}&=\underset{j}{\text{argmax}}(\mathcal{P}[j]),
\end{flalign*}
where $W_{\mathit{softmax}}\in \mathbb{R}^{4\times D}$,
$b_{\mathit{softmax}}\in \mathbb{R}^4$, $\mathcal{P}\in \mathbb{R}^4$, $j=$
class value ($0$ or $1$ or $2$ or $3$), and $\hat{y}=$ estimated class value.

\begin{table}[t]
    \centering
        \caption{Class-wise accuracy and f-score for IEMOCAP dataset for trimodal scenario.}
    \begin{tabular}[t]{ccccc}
      \hline
      \multirow{2}{*}{Metrics} & \multicolumn{4}{c}{Classes}\\
      \cline{2-5} & Happy & Sad & Neutral & Anger\\
      \hline
      Accuracy & 74.3 & 75.6 & 78.4 & 79.6 \\
      F-Score & 81.4 & 77.0 & 71.2 & 77.6 \\
      \hline
    \end{tabular}
    \label{table:iemocap-classwise}
\end{table}
Here as well, we achieve performance improvement consistent with CMU-MOSI. This
method performs 1--2.4\% better than the state of the art for all the modality
combinations. Also, trimodal accuracy is 3\% higher than the same for textual
modality. Since, IEMOCAP dataset imbalanced, we also present the f-score for each modality combination for a better evaluation. One key observation for IEMOCAP dataset is that its A+V modality
combination performs significantly better than the same of CMU-MOSI dataset. We
think that this is due to the audio and video modality of IEMOCAP being richer than
the same of CMU-MOSI. The performance difference with another strong baseline~\citep{zadten} is even more ranging from 2.1\% to 3\% on CMU-MOSI dataset and 2.2\% to 5\% on IEMOCAP dataset. This again confirms the superiority of the hierarchical fusion in compare to~\citep{zadten}. We think this is mainly because of learning the weights of bimodal and trimodal correlation (representing the degree of correlations) calculations at the time of fusion while Tensor Fusion Network (TFN) just relies on the non-trainable outer product of tensors to model such correlations for fusion.
Additionally, we present class-wise accuracy and f-score for IEMOCAP for trimodal (A+V+T) scenario in \cref{table:iemocap-classwise}.

\subsubsection{HFusion vs.\ CHFusion}

We compare HFusion and CHFusion models over CMU-MOSI dataset. We observe that
CHFusion performs 1--2\% better than HFusion model for all the modality
combinations. This performance boost is achieved by the inclusion of
utterance-level contextual information in HFusion model by adding GRUs in
different levels of fusion hierarchy.

\section{Conclusion}
\label{sec:conclusions}
Multimodal fusion strategy is an important issue in multimodal sentiment analysis. 
However, little work has been done so far in this direction. 
In this paper, we have presented a novel and comprehensive fusion strategy. 
Our method outperforms the widely used early fusion on both datasets typically used to test multimodal sentiment analysis methods.
Moreover, with the addition of context modeling with GRU, 
our method outperforms the state of the art in multimodal sentiment analysis and emotion detection by significant margin. 

In our future work, we plan to improve the quality of unimodal features, especially textual features, which will further improve the accuracy of classification.
We will also experiment with more sophisticated network architectures.

\section*{Acknowledgement}
The work was partially supported by the Instituto Polit\'ecnico Nacional via grant SIP 20172008 to A.~Gelbukh.

\bibliographystyle{elsarticle-num-names}
\bibliography{bibexport_new}
\end{document}